\documentclass{article}

\usepackage[final,nonatbib]{neurips_2021}

\usepackage[utf8]{inputenc} 
\usepackage[T1]{fontenc}    
\usepackage{hyperref}       
\usepackage{url}            
\usepackage{booktabs}       
\usepackage{amsfonts}       
\usepackage{nicefrac}       
\usepackage{microtype}      
\usepackage{xcolor}         

\usepackage{booktabs,subcaption,amsfonts,dcolumn}

\usepackage{amsmath}
\usepackage{graphicx}
\usepackage{amsmath,amssymb,booktabs,tabulary,multirow,overpic,xcolor}
\usepackage{dsfont}

\usepackage{wrapfig}

\usepackage{float}
\floatstyle{plaintop}
\restylefloat{table}

\DeclareMathOperator*{\argmin}{arg\,min}

\definecolor{pearThree}{HTML}{E74C3C}
\definecolor{pearDark}{HTML}{2980B9}
\definecolor{pearDarker}{HTML}{1D2DEC}

\hypersetup{
	colorlinks,
	citecolor=pearDark,
	linkcolor=pearThree,
    breaklinks=true,
	urlcolor=pearDarker}

\newcommand{\eg}{\emph{e.g.}}
\newcommand{\ie}{\emph{i.e.}}

\title{
Direct Multi-view Multi-person 3D  Pose Estimation
}

\newcommand{\nameofmodel}[1]{MvP}
\newcommand{\nameofattention}[1]{Projective Attention}

\makeatletter
\newcommand{\specificthanks}[1]{\@fnsymbol{#1}}
\makeatother

\author{%
  Tao Wang$^{1,2}$\thanks{Equal Contribution.},
  Jianfeng Zhang$^{2*}$,
  Yujun Cai$^1$,
  Shuicheng Yan$^1$,
  Jiashi Feng$^1$,
  \\
  $^1$Sea AI Lab
  $^2$National University of Singapore, \\
  \texttt{twangnh@gmail.com}, \\
  \texttt{zhangjianfeng@u.nus.edu},\\
  \texttt{\{caiyj,yansc,fengjs\}@sea.com}
}
  
\begin{document}

\maketitle

\begin{abstract}

We present \textbf{M}ulti-\textbf{v}iew \textbf{P}ose transformer (MvP) for estimating multi-person 3D poses from multi-view images. Instead of estimating 3D joint locations from costly volumetric representation or reconstructing the per-person 3D pose from multiple detected 2D poses as in previous methods, MvP directly regresses the multi-person 3D poses in a clean and efficient way, without relying on intermediate tasks.
Specifically,  MvP 
represents skeleton joints as learnable query
embeddings and let them progressively attend to and reason over the multi-view information from the input images to directly regress the actual 3D joint locations. 
To improve the accuracy of such a simple pipeline, MvP presents a hierarchical scheme to concisely represent query embeddings of multi-person skeleton joints and introduces an input-dependent query adaptation approach. 
Further, MvP designs a novel geometrically guided attention mechanism, called \textit{projective attention},  to more precisely fuse the cross-view information for each joint. MvP also introduces a RayConv operation to integrate the view-dependent camera geometry into the feature representations for augmenting the projective attention. 
We show experimentally that our MvP model outperforms the state-of-the-art methods on several benchmarks while being much more efficient.
Notably, it achieves 92.3\% AP$_{25}$ on the challenging Panoptic dataset, improving upon the previous best approach~\cite{Tu2020} by 9.8\%.
\nameofmodel{} is general and also extendable to recovering human mesh represented by the SMPL model, thus useful for modeling multi-person body shapes. Code and models are available at \url{https://github.com/sail-sg/mvp}.

\end{abstract}

\section{Introduction}

\begin{figure}[t]
	\centering
	\includegraphics[width=\linewidth]{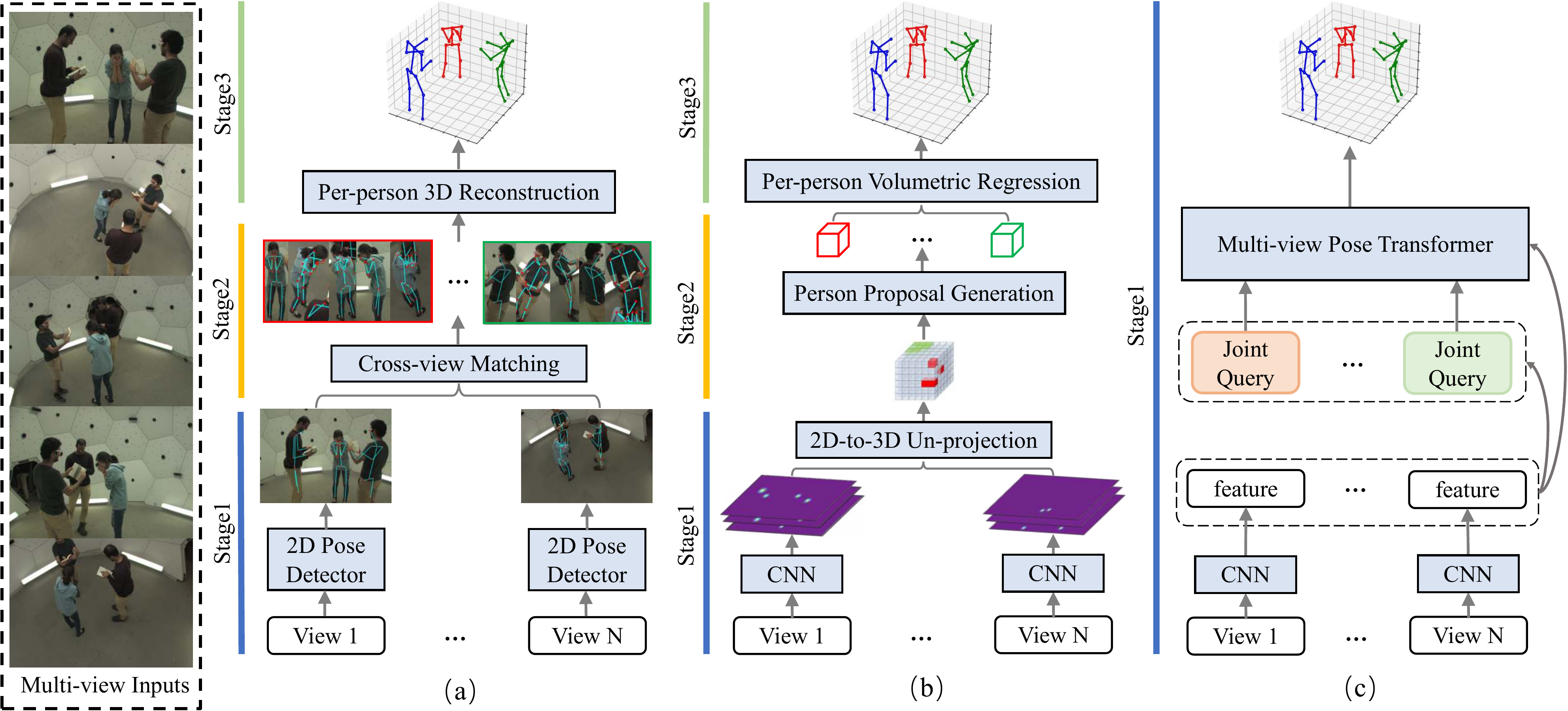}
	\caption{Difference between our method and others for multi-view multi-person 3D pose estimation. Existing methods adopt  complex multi-stage pipelines that are either (a) reconstruction-based or (b) volumetric representation based, which incur heavy computation burden. (c) Our method solves this task as a \textbf{direct regression problem} without relying on any intermediate task by a novel Multi-view Pose Transformer, and largely simplifies the pipeline and boosts the efficiency.}
	\label{fig:compare_to_others}
\end{figure}

Multi-view multi-person 3D pose estimation aims to localize 3D skeleton joints for each person instance in a scene from multi-view camera inputs. 
It is a fundamental task that benefits many real-world applications (such as surveillance, sportscast, gaming and mixed reality) and 
is mainly tackled by reconstruction-based~\cite{dong2019fast,huang2020end,chen2020multi} and volumetric~\cite{Tu2020} approaches in previous literature, as shown in Fig.~\ref{fig:compare_to_others} (a) and (b).
The former first estimates 2D poses in each view independently 
and then aggregates them and reconstructs their 3D counterparts via triangulation or a 3D pictorial structure model.
The volumetric approach~\cite{Tu2020} builds a 3D feature volume through heatmap estimation and 2D-to-3D un-projection at first, based on which instance localization and 3D pose estimation are performed for each person instance individually.
Though with notable accuracy, the above paradigms are inefficient 
due to highly relying on those intermediate tasks.
Moreover, they estimate 3D pose for each person separately, making the computation cost grow linearly with the number of persons. 

Targeted at a more simplified and efficient pipeline, we were wondering if it is possible to \textit{directly} regress 3D poses from multi-view images  without relying on any intermediate task? 
Though conceptually attractive, adopting such a direct mapping paradigm is highly non-trivial as it remains unclear how to perform skeleton joints detection and association for multiple persons within a single stage.
{In this work, we address these challenges by developing a novel \textbf{M}ulti-\textbf{v}iew \textbf{P}ose transformer (MvP) model which significantly  simplifies the multi-person 3D pose estimation.}
Specifically, MvP represents each skeleton joint as a learnable positional embedding, named \textit{joint query}, which is fed into the model and mapped into final 3D pose estimation directly (Fig.~\ref{fig:compare_to_others} (c)),  via a specifically designed attention mechanism to  fuse multi-view information 
and globally reason over the joint predictions to assign them to the corresponding person instances.  
We develop a  novel hierarchical query embedding scheme to represent the multi-person joint queries. It shares joint  embedding across different persons and introduces person-level query embedding to 
help the model in learning both   person-level and joint-level priors. Benefiting from  exploiting the person-joint relation,   the model can more accurately localize  the 3D joints. 
Further, we propose to update the joint queries with input-dependent scene-level information (\ie, globally pooled image features from multi-view inputs) such that the learnt joint queries can adapt to the target scene with better generalization performance. 

To effectively fuse the multi-view information, we propose a geometrically-guided projective attention mechanism. Instead of applying full attention to densely aggregate features {across spaces and views},   
{it projects the estimated 3D  joint into 2D  anchor points for different views, and then selectively fuses the multi-view local features near to these anchors   to precisely refine the 3D joint location.}
we propose to encode the camera rays into the multi-view feature representations via a novel RayConv operation to integrate multi-view positional information into the projective attention. 
In this way, the strong multi-view geometrical priors can be exploited by projective attention to obtain more accurate 3D pose estimation.

Comprehensive experiments on 3D pose benchmarks Panoptic~\cite{joo2015panoptic}, as well as Shelf and Campus~\cite{belagiannis20143d} demonstrate our MvP works very well. 
{Notably, it obtains 92.3\% AP$_{25}$ on the challenging Panoptic dataset, improving upon the previous best approach VoxelPose~\cite{Tu2020} by 9.8\%, while achieving nearly $2\times$ speed up.}
Moreover, the design ethos of our MvP can be easily extended to more complex tasks\textemdash we show that a simple body mesh branch with SMPL representation~\cite{loper2015smpl} trained on top of a pre-trained MvP can achieve competitively qualitative results.

Our contributions are summarized as follows:
1) We strive for simplicity in addressing the challenging multi-view multi-person 3D pose estimation problem
by casting it as a \textbf{direct regression problem} and accordingly develop a novel Multi-view Pose transformer (MvP) model, which achieves state-of-the-art results on the challenging Panoptic benchmark.
2) Different from   query embedding designs in most transformer models, we propose a more tailored    and concise  hierarchical joint query embedding scheme to enable  the model to effectively encode person-joint relation. 
Additionally, we mitigate the commonly faced generalization issue by a simple query adaptation strategy.
3) We propose a novel projective attention module along with a RayConv operation for fusing multi-view information effectively, which we believe are also inspiring for model designs in other multi-view 3D tasks. 

\section{Related Works}

\paragraph{3D Human Pose Estimation}
3D pose estimation from monocular inputs~\cite{martinez2017simple,mehta2017vnect,zhou2017towards,popa2017deep,sun2018integral,nie2019spm,zhang2020inference,gong2021poseaug,zhang2021bmp} is an ill-posed problem as multiple 3D predictions may result in the same 2D projection. 
To alleviate such projective ambiguities, multi-view methods have been explored.
Research works on single-person scenes use either multi-view geometry~\cite{Hartley2003MVG} for feature fusion~\cite{qiu2019cross,he2020epipolar} and triangulation~\cite{iskakov2019learnable,remelli2020lightweight}, or pictorial structure models for fast and robust 3D pose reconstruction~\cite{pavlakos2017harvesting,qiu2019cross}, achieving promising results. 
However, it is more challenging as we progress towards multi-person scenes.
Current approaches mainly exploit a multi-stage pipeline for multi-person tasks, including reconstruction-based~\cite{dong2019fast,chen2020multi,huang2020end,kadkhodamohammadi2021generalizable,lin2021multi} and volumetric~\cite{Tu2020} paradigms. 
Despite their notable accuracy, these methods suffer expensive computation cost from the intermediate tasks, such as cross-view matching and heatmap back-projection.
Moreover, the total computation cost grows linearly with the number of persons in the scene, making them hardly scalable for larger scenes.
Different from all previous approaches that rely on a multi-stage pipeline with computation redundancy, our method views multi-person 3D pose estimation as a \textbf{direct regression problem} based on a novel Multi-view Pose transformer model, enables an intermediate task-free single stage solution.

\paragraph{Attention and Transformers}
Driven by the recent success in natural language fields, there have been growing interests in exploring the Transformers for computer vision tasks, such as image recognition~\cite{dosovitskiy2020} and generation~\cite{jiang2021transgan}, as well as more complicated object detection~\cite{carion2020end,zhu2020deformabledetr} and video instance segmentation~\cite{wang2020end}.
However, multi-person 3D pose estimation has not been explored along this direction. 
In this study, we propose a novel Multi-view Pose Transformer architecture with a joint query embedding scheme and a projective attention module to regress 3D skeleton joints from multi-view images directly, delivering a simplified and effective pipeline.

\section{Multi-view Pose Transformer (MvP)}

To build a direct multi-person 3D pose estimation framework from multi-view images, we introduce a novel \textbf{M}ulti-\textbf{v}iew \textbf{P}ose transformer (MvP). MvP takes in the multi-view feature representations, and transforms them into groups of 3D joint locations directly (Fig.~\ref{fig:overview_and_pattn} (a)), delivering multi-person 3D pose results, with the following carefully designed query embedding and attention schemes for detecting and grouping the skeleton joints.

\subsection{Joint Query Embedding Scheme}
\label{sec:joint_query}

\begin{figure}[t]
	\centering
	\includegraphics[width=\linewidth]{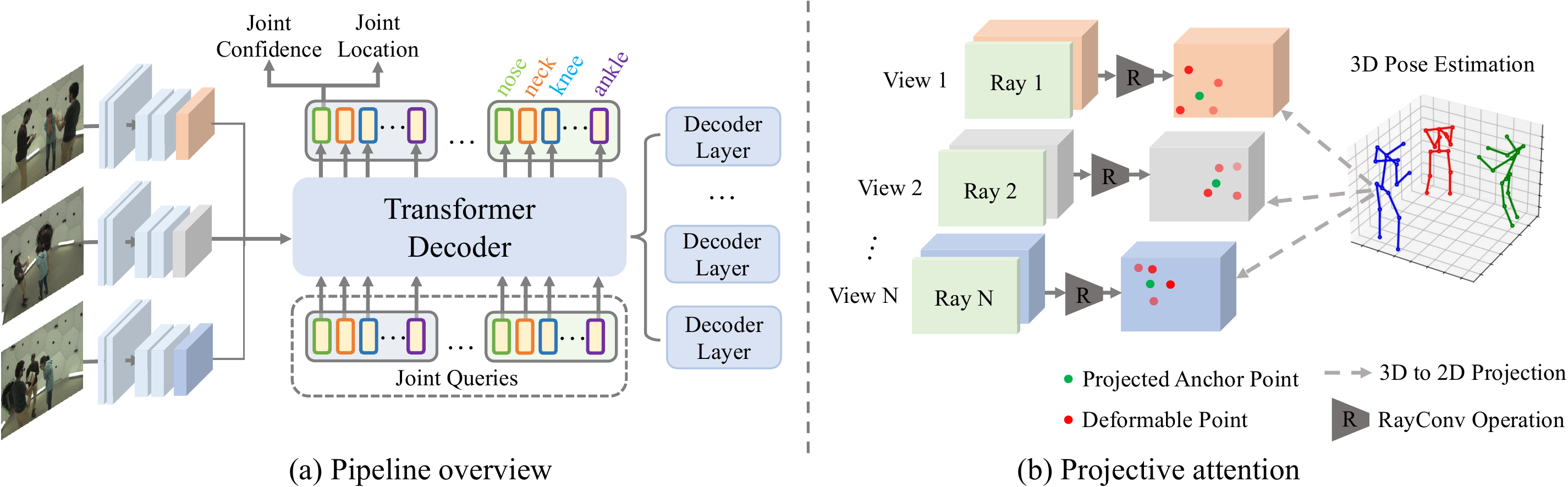}
	\caption{
	(a)  Overview of the proposed MvP model. 
	Upon the multi-view image features from several convolution layers, it deploys  a transformer decoder with a stack of decoder layers to map the input joint queries and the multi-view features to 3D poses directly.
	(b) The projective attention of MvP projects 3D skeleton joints to anchor points (the green dots) on different views and samples deformable points (the red dots) surrounding these anchors  to aggregate local contextual features via learned   weights (the brighter color density means larger weights).
	}
	\label{fig:overview_and_pattn}
\end{figure}

Inspired by transformers~\cite{vaswani2017attention}, MvP represents each skeleton joint as a learnable positional embedding, which is fed into the transformer decoder and mapped into final 3D joint location by jointly attending to other joints and the multi-view information (Fig.~\ref{fig:overview_and_pattn} (a)).
The learnt embeddings encode \emph{a prior} knowledge about the skeleton joints and we name them as \emph{joint queries}. MvP develops the following concise query embedding scheme. 

\paragraph{Hierarchical Query Embeddings}

The most straightforward way for designing joint query embeddings is to maintain a learnable query vector for each joint per person. However, we empirically find this scheme does not work well, likely because such a naive strategy cannot share the joint-level knowledge between different persons.

To tackle this problem, we develop a hierarchical query embedding scheme to explicitly encode the person-joint relation for better generalization to different scenes. 
The hierarchical embedding offers joint-level information sharing across different persons 
and reduces the learnable parameters, helping the model to learn useful knowledge from the training data, and thus generalize better.
Concretely, instead of using the set of independent joint queries $\{\textbf{q}_{m}\}^{M}_{m=1} \subset \mathbb{R}^C$, we employ a set of person level queries $\{\textbf{h}_{n}\}^{N}_{n=1} \subset \mathbb{R}^C$, and a set of joint level queries $\{\textbf{l}_{j}\}^{J}_{j=1} \subset \mathbb{R}^C$ to represent different persons and different skeleton joints, where $C$ denotes the feature dimension, $N$ is the number of persons,  $J$ is the number of joints per person, and $M=NJ$.
Then the query of joint $j$ of person $n$ can be hierarchically formulated as
\begin{equation}
	\textbf{q}_{n}^{j}=\textbf{h}_{n}+\textbf{l}_{j}.
	\label{eqn:query_add}
\end{equation}
With such a hierarchical embedding scheme, the number of learnable query embedding parameters is reduced from $NJC$ to $(N+J)C$.

\paragraph{Input-dependent Query Adaptation}
In the above, the learned joint query embeddings are shared for all the input images, independent of their   contents, and thus may not generalize well on the novel target data. 
{To address this limitation, we propose to augment the joint queries with input-dependent scene-level information in both model training and deployment, such that the learnt joint queries can be adaptive to the target data and generalize better.}
Concretely, we augment the above joint queries with a globally pooled feature vector $\textbf{g}\in \mathbb{R}^C$ from the multi-view image feature representations:
\begin{equation}
    \begin{split}
        \textbf{q}_{n}^{j}&=\textbf{g}+\textbf{h}_{n}+\textbf{l}_{j}.
    \end{split}
\end{equation}
Here $\textbf{g}=\mathrm{Concat}(\mathrm{Pool}(\textbf{Z}_1),\ldots, \mathrm{Pool}(\textbf{Z}_V))\textbf{W}^g$, where $\textbf{Z}_v$ denotes image feature from $v$-th view and $V$ is the total number of camera views; $\mathrm{Concat}$ and $\mathrm{Pool}$ denote concatenation and pooling operations, and $\textbf{W}^g$ is a learnable linear weight. 

\subsection{Projective Attention for Multi-view Feature Fusion}
\label{sec:projective_attn}
It is crucial to aggregate complementary multi-view information to transform the joint embeddings into accurate 3D joint locations.
We consider  the    dot product attention mechanism of transformers~\cite{vaswani2017attention} to fuse the multi-view image features. 
However, naively applying  such dot product attention densely over all spatial locations and camera views will incur enormous computation cost. 
Moreover, such dense attention is difficult to optimize and delivers poor performance empirically since it does not exploit any 3D geometric knowledge.

Therefore, we propose a geometrically-guided multi-view projective attention scheme, named projective attention.
{The core idea is to take the 2D projection of the estimated 3D joint location as the anchor point in each view, and only fuse the local features near those projected 2D locations from different views.}
Motivated by the deformable convolution~\cite{dai2017deformable,zhu2019deformable}, we adopt an adaptive deformable sampling strategy to gather the localized context information in each camera view, as shown in Fig.~\ref{fig:overview_and_pattn} (b). Other local attention operations~\cite{zhao2020exploring,wu2020lite,wu2019pay} can also be adopted as an alternative. Formally, given joint query feature $\textbf{q}$ and 3D joint position $\textbf{y}$, the projective attention is defined as
\begin{equation}
	\begin{split}
		\mathrm{PAttention}(\textbf{q},\textbf{y}, \{\textbf{Z}_{v}\}^{V}_{v=1})&=\mathrm{Concat}(\textbf{f}_1, \textbf{f}_2,\ldots, \textbf{f}_V)\textbf{W}^P, \\
		\text{where}~\textbf{f}_v&=\sum_{k=1}^K \textbf{a}(k) \cdot \textbf{Z}_v \big(\Pi(\textbf{y},\textbf{C}_v)+\Delta \textbf{p}(k)\big)\textbf{W}^f.
	\end{split}
	\label{eqn:p_attention}
\end{equation}
Here the view-specific feature $\textbf{f}_v$ is obtained by aggregating features from $K$ discrete offsetted sampling points from an anchor point $\textbf{p}=\Pi(\textbf{y},\textbf{C}_v)$, located by projecting the current 3D joint location $\textbf{y}$ to 2D, where $\Pi: \mathbb{R}^3 \to \mathbb{R}^2$ denotes perspective projection~\cite{Hartley2003MVG} and $\textbf{C}_v$ the corresponding camera parameters.
$\textbf{W}^P$ and $\textbf{W}^f$ are learnable linear weights. 
The attention weight $\textbf{a}$ and the offset to the projected anchor point $\Delta \textbf{p}$ are 
estimated from
the fusion of query feature $\textbf{q}$ and the view-dependent feature at the projected anchor point $\textbf{Z}_v(\textbf{p})$, \ie, $\textbf{a}=\mathrm{Softmax}((\textbf{q}+\textbf{Z}_v(\textbf{p}))\textbf{W}^a)$ and $\Delta \textbf{p}=(\textbf{q}+\textbf{Z}_v(\textbf{p}))\textbf{W}^p$, where $\textbf{W}^a$ and $\textbf{W}^p$ are learnable linear weights. 
If the projected location and the offset are fractional, we use bilinear interpolation to obtain the corresponding feature $\textbf{Z}_v (\textbf{p})$ or $\textbf{Z}_v (\textbf{p}+\Delta \textbf{p}(t))$. 

The projective attention incorporates two geometrical cues, \ie, the corresponding 2D spatial locations across views from   the 3D to 2D projection   and   the deformed neighborhood of the anchors from the learned offsets to gather view-adaptive contextual information.  
Unlike naive attention where the query feature densely interacts with the multi-view key features across all the spatial locations, 
{the projective attention is more selective for the interaction between the query and each view\textemdash only the features from locations near to the projected anchors are aggregated, and thus is much more efficient.}

\paragraph{Encoding Multi-view Positional Information with RayConv} 
The positional encoding~\cite{vaswani2017attention} is an important component of the transformer, which provides positional information of the input sequence.
However, a simple per-view 2D positional encoding scheme cannot encode the multi-view geometrical information. 
To tackle this limitation,
we propose to encode the camera ray directions {that represent  positional information in 3D space}  into the multi-view feature representations. 
Concretely, the camera ray direction $\textbf{R}_v$, generated with the view-specific camera parameters, 
is concatenated channel-wisely to the corresponding image feature representation $\textbf{Z}_v$. Then a standard convolution is applied to obtain the updated feature representation $\hat{\textbf{Z}}_v$,    with  the    view-dependent geometric information:
\begin{equation}
	\hat{\textbf{Z}}_v=\mathrm{Conv}(\mathrm{Concat}(\textbf{Z}_v, \textbf{R}_v)).
\end{equation}
{We name the operation as \emph{RayConv}. With it, the obtained feature representation $\hat{\textbf{Z}}_v$ is used for the projective attention by replacing ${\textbf{Z}}_v$ in Eqn.~\eqref{eqn:p_attention}.
 
Such drop-in replacement  
introduces negligible computation, while injecting strong multi-view geometrical prior to augment the projective attention scheme,  
thus helping more precisely predict the refined 3D joint position.
}

\subsection{Architecture}
\label{sec:arch}
Our overall architecture (Fig.~\ref{fig:overview_and_pattn} (a)) is pleasantly simple.
It adopts a convolution neural network, designed for 2D pose estimation~\cite{xiao2018simple}, to obtain high-resolution image features $\{\textbf{Z}_{v}\}^{V}_{v=1}$ from multi-view inputs $\{\textbf{I}_{v}\}^{V}_{v=1}$.
The features are then fed into the transformer decoder  consisting of   multiple decoder layers to predict the 3D joint locations.
Each layer conducts a self-attention  to perform  pair-wise   interaction between  all the joints from all the persons in the scene; a projective attention   to selectively gather the complementary multi-view information; {and a feed-forward   regression to predict the  3D joint positions and      their   confidence scores.
Specifically, the transformer decoder   applies  a \textit{multi-layer progressive regression scheme},  \ie, each decoder layer outputs 3D joint offsets to refine the input 3D joint positions from previous layer.}

\paragraph{Extending to Body Mesh Recovery}
MvP learns skeleton joints feature representations and is extendable to recovering human mesh with a parametric body mesh model~\cite{loper2015smpl}. Specifically, after average pooling on the joint features into per-person feature, a feed-forward network is used to predict the corresponding body mesh represented by the parametric SMPL model~\cite{loper2015smpl}.
Similar to the joint location prediction, the SMPL parameters follow multi-layer progressive regression scheme.

\subsection{Training}
\label{sec:training}
MvP infers a fixed set of $M$ joint locations for $N$ different persons, {where $M=NJ$}. 
The main training challenge   is how to associate the skeleton joints correctly for different person instances. Unlike the post-hoc grouping of detected skeleton joints as in bottom-up pose estimation methods~\cite{papandreou2018personlab,kreiss2019pifpaf}, MvP learns to directly predict the multi-joint 3D human pose in a group-wise fashion as shown in Fig.~\ref{fig:overview_and_pattn} (a). This is achieved by a grouped matching strategy during model training.

\paragraph{Grouped Matching}
Given the predicted joint positions $\{\textbf{y}_m\}^{M}_{m=1} \subset \mathbb{R}^{3}$ and associated confidence scores $\{s_m\} ^{M}_{m=1}$, \
{we group every consecutive $J$-joint predictions into per-person pose estimation $\{\textbf{Y}_n\} ^{N}_{n=1} \subset \mathbb{R}^{J\times 3}$, and average their corresponding confidence scores to obtain the per-person confidence scores $\{p_n\} ^{N}_{n=1}$.} The same grouping strategy is used during inference.

The ground truth set $\textbf{Y}^*$ of 3D poses of different person instances  
is smaller than the prediction set of size $N$, which is padded to size $N$ with empty element $\varnothing$. Then we find a bipartite matching between the prediction set and the ground truth set by searching for a permutation of $\hat{\sigma} \in \aleph_N$ that achieves the lowest matching cost:
\begin{equation}
	\hat{\sigma}=\argmin_{\sigma \in \aleph_N }\sum_{n=1}^N \mathcal{L}_\mathrm{match}(\textbf{Y}^*_n,\textbf{Y}_{\sigma(n)}).
\end{equation}
We consider both the regressed 3D joint position and confidence score for the matching cost:
\begin{equation}
	\mathcal{L}_\mathrm{match}(\textbf{Y}^*_n,\textbf{Y}_{\sigma(n)})= -p_i+\mathcal{L}_1(\textbf{Y}^*_n,\textbf{Y}_{\sigma(n)})
\end{equation}
where $\textbf{Y}^*_n\neq\varnothing$, and $\mathcal{L}_1$ computes the $L_1$ loss error. 
Following~\cite{carion2020end,sutskever2014sequence}, we employ the Hungarian algorithm~\cite{kuhn1955hungarian} to compute the optimal assignment $\hat{\sigma}$ with the above matching cost.

\paragraph{Objective Function}
We compute the \textit{Hungarian loss} with the obtained optimal assignment $\hat{\sigma}$: 
\begin{equation}
	\mathcal{L}_\mathrm{Hungarian}(\textbf{Y}^*,\textbf{Y})=\sum_{n=1}^N \left[  \mathcal{L}_\mathrm{conf}(\textbf{Y}^*_n, p_{\hat{\sigma}(n)}) + \mathds{1}_{\{\textbf{Y}^*_n\neq \varnothing\}}\lambda \mathcal{L}_\mathrm{pose}(\textbf{Y}^*_n,\textbf{Y}_{\hat{\sigma}(n)})  \right].
\end{equation}
Here $\mathcal{L}_\mathrm{conf}$ and $\mathcal{L}_\mathrm{pose}$ are losses for confidence score and pose regression, respectively. $\lambda$ balances the two loss terms. We use focal loss~\cite{lin2017focal} for confidence prediction which adaptively balances the positive and negative samples.
{For pose regression, we compute $L_1$ loss for 3D joints and their projected 2D joints in different views. 
}

{To learn multi-layer progressive regression, the above matching and loss are applied for each decoder layer. The total loss is thus $\mathcal{L}_\mathrm{total}=\sum_{l=1}^L \mathcal{L}^{l}_\mathrm{Hungarian}$, where $\mathcal{L}^{l}_\mathrm{Hungarian}$ denotes loss of the $l$-th decoder layer and $L$ is the number of decoder layers. When extending MvP to body mesh recovery, we apply $L_1$ loss for 3D joints from the SMPL model and their 2D projections, as well as an adversarial loss following HMR~\cite{hmrKanazawa17,jiang2020coherent,zhang2021bmp} due to lack of GT SMPL parameters.}

\section{Experiments}
\label{experiments}
In this section, we aim to answer following questions.
1) Can MvP provide both efficient and accurate 
multi-person 3D pose estimation?
2) How does the proposed attention mechanism help multi-view multi-person skeleton joints information fusing?
3) How does each individual design choice affect model performance?
To this end, we conduct extensive experiments on several benchmark datasets.

\begin{table}[b]
	\renewcommand{\tabcolsep}{4pt}
	\centering
	\small
	\caption{Result on the Panoptic dataset. MvP is more accurate and faster than VoxelPose. }
	\begin{tabular}{cccccccc}
	\toprule
	Methods & AP$_{25}$ & AP$_{50}$ & AP$_{100}$ &  AP$_{150}$ & Recall$_{@500}$ & MPJPE[mm] & Time[ms] \\
	\midrule
	VoxelPose~\cite{Tu2020} & 84.0 & 96.4 & \textbf{97.5} & \textbf{97.8} & 98.1 & 17.8 & 320\\
	MvP (Ours) &\textbf{92.3} & \textbf{96.6} & \textbf{97.5} & 97.7 & \textbf{98.2} & \textbf{15.8} & \textbf{170}\\
	\bottomrule
	\end{tabular}
	\label{panoptic_main}
\end{table}

\paragraph{Datasets}
\textit{Panoptic}~\cite{joo2017panoptic}  is a large-scale benchmark with 3D skeleton joint annotations. It captures daily social activities in an indoor environment. 
We conduct extensive experiments on Panoptic to evaluate and analyze our approach.
Following VoxelPose~\cite{Tu2020}, we use the same data sequences except `160906\_band3' in the training set due to broken images. Unless otherwise stated, we use five HD cameras (3, 6, 12, 13, 23) in our experiments. All results reported in the experiments follow the same data setup. We use Average Precision (AP) and Recall~\cite{Tu2020}, as well as Mean Per Joint Position Error (MPJPE) as evaluation metrics.
\textit{Shelf} and \textit{Campus}~\cite{belagiannis20143d}  are two multi-person datasets capturing indoor and outdoor environments, respectively. We split them into training and testing sets following~\cite{belagiannis20143d,dong2019fast,Tu2020}. We report Percentage of Correct Parts (PCP) for these two datasets.

\paragraph{Implementation Details}
Following VoxelPose~\cite{Tu2020}, we adopt a pose estimation model~\cite{xiao2018simple} build upon ResNet-50~\cite{he2016deep} for multi-view image features extraction. 
Unless otherwise stated, we use a stack of six transformer decoder layers. The model is trained for 40 epochs, with the Adam optimizer of learning rate $10^{-4}$. During inference, a confidence threshold of 0.1 is used to filter out redundant predictions. Please refer to supplementary for more implementation details.

\subsection{Main Results}
\begin{wrapfigure}{r}{0.5\textwidth}
\centering
\includegraphics[width=0.5\textwidth]{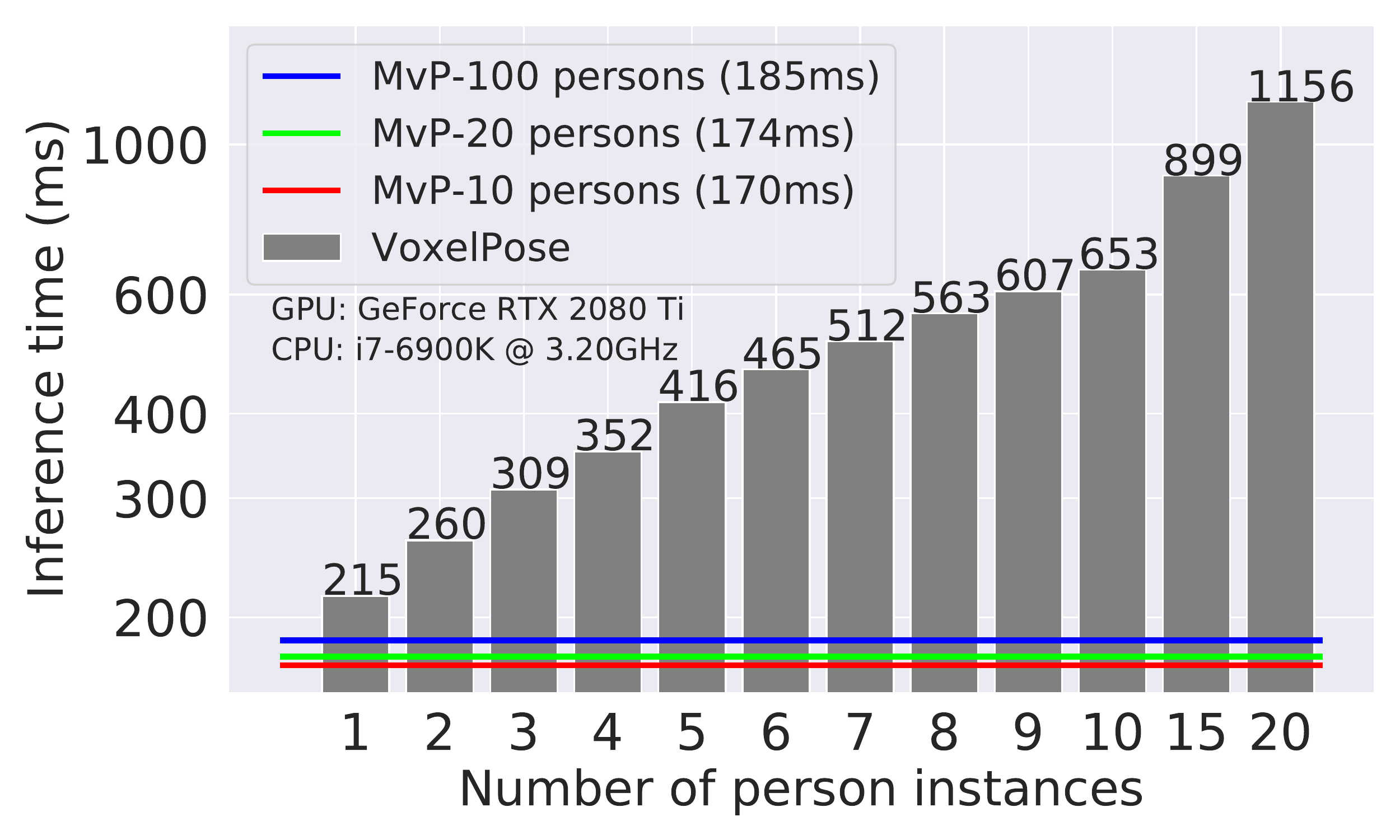}
\caption{Inference time versus the number of person instances. Benefiting from its direct inference framework, MvP maintains almost constant inference time regardless of the number of persons.}
\label{ablation:inf_time_vs_person_num}
\end{wrapfigure}
\paragraph{Panoptic} 
We first evaluate our MvP model on the challenging Panoptic dataset and compare it with the state-of-the-art VoxelPose model~\cite{Tu2020}.
As shown in Table~\ref{panoptic_main}, Our MvP achieves 92.3 AP$_{25}$, improving upon VoxelPose by 9.8\%, and achieves   much lower MPJPE (15.8 \textit{v.s} 17.8). 
Moreover,  MvP only requires 170ms to process a multi-view input, about $2\times$ faster than VoxelPose\footnote{We count averaged per-sample inference time in millisecond on Panoptic test set. For all methods, the time is counted on GPU GeForce RTX 2080 Ti and CPU Intel i7-6900K @ 3.20GHz.}. These results demonstrate both accuracy and efficiency advantages  of MvP from estimating 3D poses of multiple persons in a direct regression paradigm.
To further demonstrate   efficiency of MvP, we compare its inference time  with VoxelPose's when processing different numbers of person instances.
As shown in Fig.~\ref{ablation:inf_time_vs_person_num}, the inference time of VoxelPose grows linearly with the number of persons in the scene due to the per-person regression paradigm. In contrast, MvP keeps constant inference time no matter how many instances in the scene. Notably, it takes only 185ms for MvP to process scenes even with 100 person instances (the blue line),  demonstrating its great potential to handle crowded scenarios.

\paragraph{Shelf and Campus} 
We further compare our MvP with state-of-the-art approaches on the Shelf and Campus datasets.
The reconstruction-based methods~\cite{belagiannis20153d,ershadi2018multiple,dong2019fast} use 3D pictorial model~\cite{belagiannis20153d,dong2019fast} or conditional random field~\cite{ershadi2018multiple} within a multi-stage paradigm; and the volumetric approach VoxelPose~\cite{Tu2020} highly relies on computationally intensive intermediate tasks. As shown in Table~\ref{shelf_campus}, our MvP achieves the best performance in all the actors on the Shelf dataset. Moreover, it obtains a comparable result on the Campus dataset as VoxelPose~\cite{Tu2020} without relying on any intermediate task. These results further confirm the effectiveness of MvP for estimating 3D poses of multiple persons directly.

\begin{table}[h]
\centering
\caption{Results (in PCP) on Shelf and Campus datasets.}
\renewcommand{\tabcolsep}{3.5pt}
\small
\begin{tabular}{c|cccc|cccc}
\toprule
\multirow{2}{*}{Methods} & \multicolumn{4}{c|}{Shelf} & \multicolumn{4}{c}{Campus} \\ \cmidrule{2-9} 
        & Actor 1 & Actor 2 & Actor 3 & Average & Actor 1 & Actor 2 & Actor 3 &  Average   \\ \midrule
Belagiannis \textit{et al.}~\cite{belagiannis20153d}       &   75.3  &  69.7   &   87.6  &  77.5   &  93.5   &  75.7   & 84.4    &    84.5 \\ 
Ershadi \textit{et al.}~\cite{ershadi2018multiple}       &   93.3  &  75.9   &   94.8  &  88.0   &  94.2   &  92.9   & 84.6    &    90.6 \\ 
Dong \textit{et al.}~\cite{dong2019fast}       &   98.8  &  94.1   &  \textbf{97.8}  &  96.9   &  97.6   &  93.3   & 98.0    &    96.3 \\ 
VoxelPose~\cite{Tu2020}       &   \textbf{99.3}  & 94.1   &   97.6 &  97.0   &  97.6   &  {93.8}   & \textbf{98.8}    &    \textbf{96.7} \\ 
MvP (Ours)   &   \textbf{99.3}  & \textbf{95.1}   &   \textbf{97.8} &  \textbf{97.4}   &  \textbf{98.2}   &  \textbf{94.1}   & 97.4    &    96.6 \\ \bottomrule
\end{tabular}
\label{shelf_campus}
\end{table}

\subsection{Visualization}

\begin{figure}[t]
	\centering
	\includegraphics[width=0.9\linewidth]{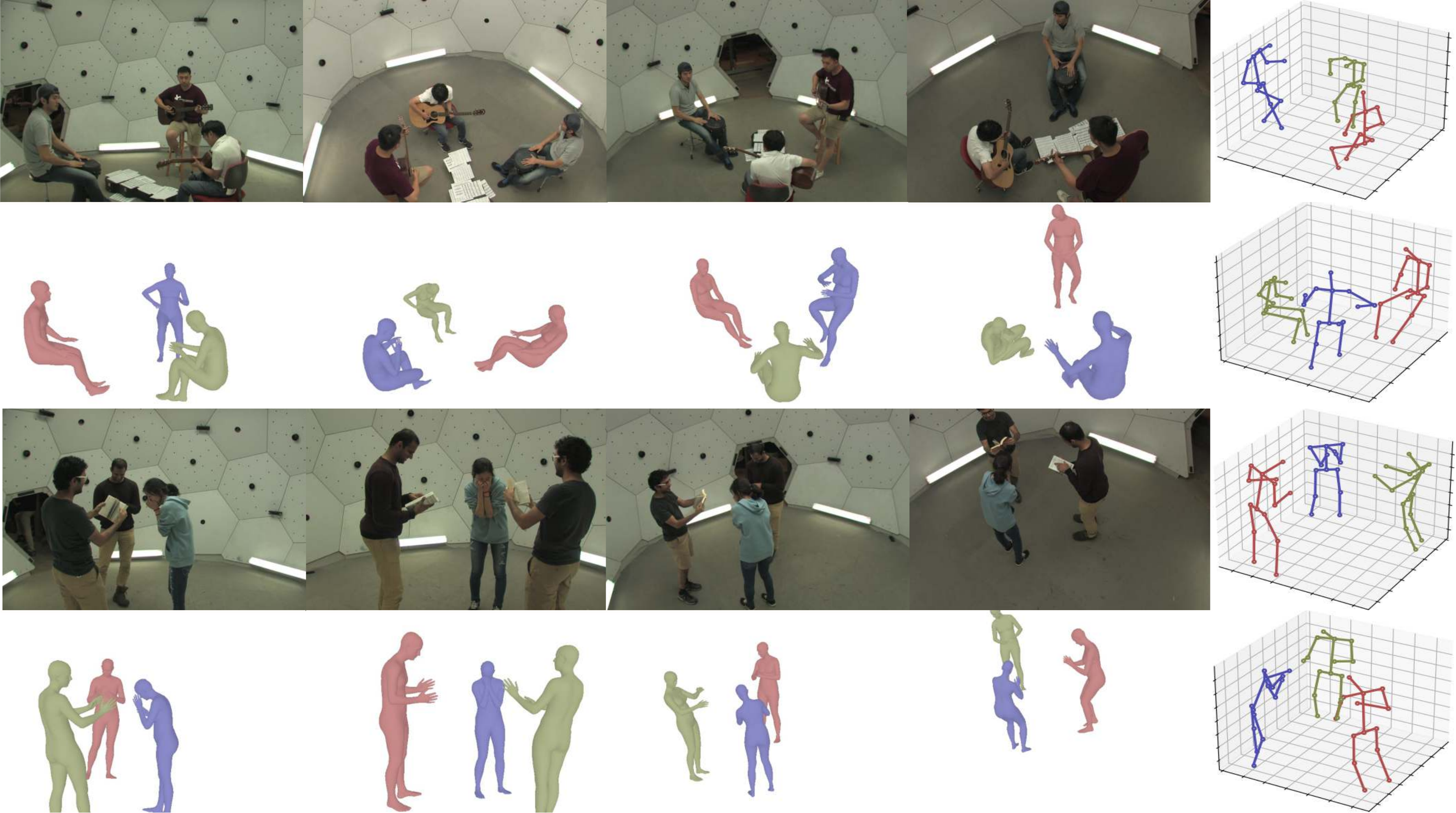}
	\caption{
    Example 3D pose estimations from Panoptic dataset. The left four columns show the multi-view inputs and the corresponding body mesh estimations. The rightmost column shows the estimated 3D poses from two different viewpoints.
    Best viewed in color. 
	}
	\label{fig:exp_vis}
\end{figure}

\paragraph{3D Pose and Body Mesh Estimation} 
We visualize some 3D pose estimations of MvP on the challenging Panoptic dataset in Fig.~\ref{fig:exp_vis}. It can be observed that MvP is robust to large pose deformation (the 1st example) and severe occlusion (the 2nd example), and can achieve geometrically plausible results w.r.t. different viewpoints (the rightmost column). Moreover, MvP is extendable to body mesh recovery and can achieve fairly good reconstruction results (the 2nd and 4th rows). All these results verify both effectiveness and extendability of MvP. 
Please see supplementary for more examples.

\begin{figure}[t]
	\centering
	\includegraphics[width=0.9\linewidth]{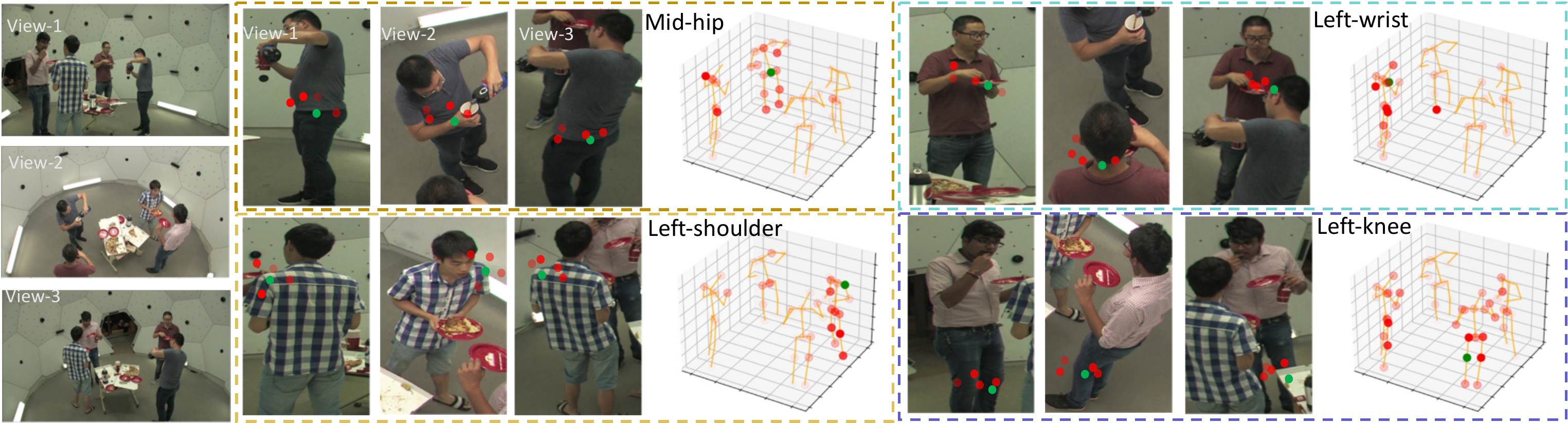}
	\caption{Visualization of projetive attention   and self-attention   on example skeleton joints. The attention weights are obtained with the $4$-th decoder layer of a trained model. 
	\textit{Projective attention} (in the cropped image triplets):  the green points denote the projected anchor points in each camera view, and the red points denote the offsetted spatial locations, with brighter color for stronger attention. 
	\textit{Self-attention} (in the 3D skeleton plots):   example skeleton joint (green) to all the other skeleton joints (red) in the scene. The color density indicates attention weight. Best viewed in color and $2\times$ zoom.}
	\label{fig:exp_vis_attention}
\end{figure}

\paragraph{Attention Mechanism} 
We visualize the projective attention and the self-attention in Fig.~\ref{fig:exp_vis_attention}. Benefiting from the 3D-to-2D projection, the projective attention can accurately locate the skeleton joint in each camera view (the green point) based on the current estimated 3D joint location. We observe it learns to gather adaptive local context information (the red points) with the deformable sampling operation. For instance, when regressing the 3D position of mid-hip (the 1st example), the projective attention selectively attends to informative joints such as the left and right hips as well as thorax, which offers sufficient contextual information 
for   accurate estimation. We also visualize the self-attention, which learns pair-wise interaction between all the skeleton joints in the scene. From the 3D plot in Fig.~\ref{fig:exp_vis_attention}, we can observe a certain skeleton joint mainly attends to other joints of the same person instance (more opaque). It also attends to joints from other person instances, but with less attention (more transparent). This phenomenon is reasonable as the skeleton joints of a human body are strongly correlated to each other, \eg, with certain pose priors and bone length.

\subsection{Ablation}

\begin{table}[t]
	\renewcommand{\tabcolsep}{2pt}
	\small
	\begin{subtable}[!t]{0.32\textwidth}
		\centering
		\begin{tabular}{cccccccc}
			\toprule
			\textit{RConv} & AP$_{25}$ & AP$_{100}$ & MPJPE \\
			\midrule
			w/  & 92.3 & 97.5 & 15.8\\
			\ \ w/o  & 87.5 & 96.2 & 17.4\\
			\bottomrule
		\end{tabular}
		\caption{The effect of \emph{RayConv}. w/o means removing RayConv.
		}
		\label{ablation:rayconv}
	\end{subtable}
	\hspace{\fill}
	\begin{subtable}[!t]{0.32\textwidth}
		\centering
		\begin{tabular}{cccccccc}
			\toprule
			\textit{Query} & AP$_{25}$ & AP$_{100}$ & MPJPE \\
			\midrule
			Per-joint & 67.4 & 84.7 & 41.2\\
			Hier. & 82.5 & 93.2 & 19.5\\
			Hier.+ad. & 92.3 & 97.5 & 15.8\\
			\bottomrule
		\end{tabular}
		\caption{Different joint query embedding schemes. 
		}
		\label{ablation:query_embed}
	\end{subtable}
	\hspace{\fill} 
	\begin{subtable}[!t]{0.32\textwidth}
		\centering
		\begin{tabular}{cccccccc}
			\toprule
			\textit{Thr.} & AP$_{25}$ & AP$_{100}$ &  MPJPE \\
			\midrule
			0.0 & 93.1 & 98.5 & 16.3\\
			0.1 & 92.3 & 97.5 & 15.8\\
			0.2 & 91.1 & 96.2 & 15.5\\
			0.4 & 89.2 & 93.7 & 15.0\\
			\bottomrule
		\end{tabular}
		\caption{Different confidence threshold during evaluation. 
		}
		\label{ablation:conf_threshold}
	\end{subtable}
	\hspace{\fill}
	\begin{subtable}[b]{0.32\textwidth}
		\centering
		\begin{tabular}{cccccccc}
			\toprule
			\textit{Dec.} & AP$_{25}$ & AP$_{100}$ & MPJPE \\
			\midrule
			2 & 6.3 & 92.5 & 49.6\\
			3 & 63.4 & 95.6 & 22.8\\
			4 & 86.8 & 96.8& 17.5\\
			5 & 91.8 & 97.6 & 16.2\\
			6 & 92.3 & 97.5 & 15.8\\
			7 & 92.0 & 97.5 & 15.9\\
			\bottomrule
		\end{tabular}
		\caption{Number of decoder layers. 
		}
		\label{ablation:decoder_layer}
	\end{subtable}
	\hspace{\fill}
	\begin{subtable}[b]{0.32\textwidth}
		\centering
		\begin{tabular}{cccccccc}
			\toprule
			\textit{Cam.} & AP$_{25}$ & AP$_{100}$ & MPJPE \\
			\midrule
			1 & 4.7 & 61.0 & 93.8\\
			2 & 37.7 & 93.0 & 34.8\\
			3 & 71.8 & 95.1 & 21.1\\
			4 & 84.1 & 96.7 & 19.3\\
			5 & 92.3 & 97.5 & 15.8\\
			\bottomrule
		\end{tabular}
		\caption{Number of camera views. 
		}
		\label{ablation:camera_view}
	\end{subtable}
	\hspace{\fill}
	\begin{subtable}[b]{0.32\textwidth}
		\centering
		\begin{tabular}{cccccccc}
			\toprule
			$K$ & AP$_{25}$ & AP$_{100}$ &  MPJPE \\
			\midrule
			1 & 88.6 & 96.3 & 18.2\\
			2 & 89.3 & 97.5 & 17.4\\
			4 & 92.3 & 97.7 & 15.8\\
			8 & 84.4 & 91.1 & 20.3\\
			\bottomrule
		\end{tabular}
		\caption{Number of deformable points $K$. 
		}
		\label{ablation:deform_points}
	\end{subtable}
	\hspace{\fill}
	\caption{Ablations on Panoptic. In (b), \textit{Hier.} denotes the hierarchical query embedding scheme, \textit{Hier.+ad.} means further adding the adaptation strategy. Please see supplement for more ablations.}
	\label{ablations}
\end{table}

\paragraph{Importance of RayConv} MvP introduces RayConv to encode multi-view geometric information, \ie, camera ray directions into image feature representations. 
As shown in Table~\ref{ablation:rayconv}, if removing RayConv, the performance drops significantly\textemdash4.8 decrease in AP$_{25}$ and 1.6 increase in MPJPE. This indicates  the multi-view geometrical information is important for the model to more precisely localize the skeleton joints in 3D space. {Without RayConv, the transformer decoder cannot accurately capture positional information in 3D space, resulting in performance drop.}

\paragraph{Importance of Hierarchical Query Embedding}
As shown in Table~\ref{ablation:query_embed}, compared with  the straightforward and unstructured per-joint query embedding scheme,  the proposed hierarchical query embedding boosts the performance sharply\textemdash 14.1 increase  in AP$_{25}$ and 23.4 decrease in MPJPE. 
Its  advantageous performance clearly  verifies  introducing the person-level queries to collaborate with the joint-level queries can better exploit human body structural information   and improve model to better localize the joints. 
 Upon the hierarchical query embedding scheme, adding the query adaptation strategy further improves the performance significantly, reaching AP$_{25}$ of 92.3 and MPJPE of 15.8. This shows the proposed approach effectively adapts the query embeddings to the target scene and such adaptation is indeed beneficial for the generalization of  MvP to novel scenes.

\paragraph{Different Model Designs } We also examine effects of varying the following    designs   of the MvP model to gain better understanding on them.

\textbf{Confidence Threshold } During inference, a confidence threshold is used to to filter out the low-confidence and erroneous pose predictions, and obtain the final result. Adopting a higher confidence will select the predictions in a more restrictive way. 
As shown in Table~\ref{ablation:conf_threshold}, a higher confidence threshold brings lower MPJPE as it selects more accurate predictions; but it also filters out some true positive predictions and thus reduces the average precision.

\textbf{Number of Decoder Layers } Decoder layers are used for refining the pose estimation.  Stacking more decoder layers thus gives better performance (Table~\ref{ablation:decoder_layer}). For instance, the MPJPE is as high as 49.6 when using only two decoder layers, but it is significantly reduced to 22.8 when using three decoder layers. This clearly justifies the progressive refinement strategy of our MvP model is effective.   However the benefit of using more decoder layers diminishes when the number of layers is large enough, implying the model has reached the ceiling of its model  capacity.

\textbf{Number of Camera Views } Multi-view inputs provide complementary information to each other which is extremely useful when 
handling some challenging environment factors in 3D pose estimation like occlusions. We vary the number of camera views to examine whether MvP can effectively fuse and leverage multi-view information to continuously improve the pose estimation quality (Table~\ref{ablation:camera_view}). 
As expected, with more camera views, the 3D pose estimation accuracy monotonically increases, demonstrating the capacity of MvP in fusing multi-view information.

\textbf{Number of Deformable Sampling Points }  Table~\ref{ablation:deform_points} shows the effect of the number of deformable sampling points $K$ used in the projective attention. With only one deformable point, MvP already achieves a respectable result, \ie, 88.6 in AP$_{25}$ and 17.4 in MPJPE. Using more sampling points further improves the performance, demonstrating the projective attention is effective at aggregating information from the useful locations.  When $K=4$, the model gives the best result. Further increasing $K$ to 8, the performance starts to drop. It is likely because using too many deformable points introduces redundant information and  thus makes the model more difficult to optimize. 

\section{Conclusion}
\label{sec:conclusion}
We introduced a direct and efficient model,  named Multi-view Pose transformer (MvP),  to address the challenging multi-view multi-person 3D human pose estimation problem. Different from existing methods relying on tedious intermediate tasks,  MvP substantially simplifies the pipeline into a direct regression one by carefully designing the transformer-alike model architecture with a novel hierarchical joint query embedding scheme and projective attention mechanism. We conducted extensive experiments to verify its superior performance and speed over the well-established baselines.  

We empirically found MvP needs sufficient data for model training since it learns the 3D geometry implicitly. In the future, we will study how to enhance the data-efficiency of MvP by leveraging the strategy like self-supervised pre-training or exploring more advanced approaches. Similar to prior works, we also found MvP suffers from performance drop for cross-camera generalization, that is, generalizing on novel camera views. We will explore approaches like disentangling camera parameters and multi-view feature learning to improve this aspect.
Besides, we will explore the large-scale applications of MvP and further extend it to other relevant tasks. Thanks to its efficiency, MvP would be scalable to handle very crowded  scenes with many persons. Moreover, the framework of MvP is general and thus  extensible to other 3D modeling tasks like dense mesh recovery of common objects. 


{\small
	\bibliographystyle{plain}
	\bibliography{egbib}

\begin{thebibliography}{10}

\bibitem{belagiannis20143d}
Vasileios Belagiannis, Sikandar Amin, Mykhaylo Andriluka, Bernt Schiele, Nassir
  Navab, and Slobodan Ilic.
\newblock 3d pictorial structures for multiple human pose estimation.
\newblock In {\em {CVPR}}, 2014.

\bibitem{belagiannis20153d}
Vasileios Belagiannis, Sikandar Amin, Mykhaylo Andriluka, Bernt Schiele, Nassir
  Navab, and Slobodan Ilic.
\newblock 3d pictorial structures revisited: Multiple human pose estimation.
\newblock {\em IEEE transactions on pattern analysis and machine intelligence},
  38(10):1929--1942, 2015.

\bibitem{carion2020end}
Nicolas Carion, Francisco Massa, Gabriel Synnaeve, Nicolas Usunier, Alexander
  Kirillov, and Sergey Zagoruyko.
\newblock End-to-end object detection with transformers.
\newblock In {\em {ECCV}}, 2020.

\bibitem{chen2020multi}
He~Chen, Pengfei Guo, Pengfei Li, Gim~Hee Lee, and Gregory Chirikjian.
\newblock Multi-person 3d pose estimation in crowded scenes based on multi-view
  geometry.
\newblock In {\em {ECCV}}, 2020.

\bibitem{dai2017deformable}
Jifeng Dai, Haozhi Qi, Yuwen Xiong, Yi~Li, Guodong Zhang, Han Hu, and Yichen
  Wei.
\newblock Deformable convolutional networks.
\newblock In {\em {ICCV}}, 2017.

\bibitem{dong2019fast}
Junting Dong, Wen Jiang, Qixing Huang, Hujun Bao, and Xiaowei Zhou.
\newblock Fast and robust multi-person 3d pose estimation from multiple views.
\newblock In {\em {CVPR}}, 2019.

\bibitem{dosovitskiy2020image}
Alexey Dosovitskiy, Lucas Beyer, Alexander Kolesnikov, Dirk Weissenborn,
  Xiaohua Zhai, Thomas Unterthiner, Mostafa Dehghani, Matthias Minderer, Georg
  Heigold, Sylvain Gelly, et~al.
\newblock An image is worth 16x16 words: Transformers for image recognition at
  scale.
\newblock {\em arXiv}, 2020.

\bibitem{dosovitskiy2020}
Alexey Dosovitskiy, Lucas Beyer, Alexander Kolesnikov, Dirk Weissenborn,
  Xiaohua Zhai, Thomas Unterthiner, Mostafa Dehghani, Matthias Minderer, Georg
  Heigold, Sylvain Gelly, Jakob Uszkoreit, and Neil Houlsby.
\newblock An image is worth 16x16 words: Transformers for image recognition at
  scale.
\newblock {\em arXiv}, 2020.

\bibitem{ershadi2018multiple}
Sara Ershadi-Nasab, Erfan Noury, Shohreh Kasaei, and Esmaeil Sanaei.
\newblock Multiple human 3d pose estimation from multiview images.
\newblock {\em Multimedia Tools and Applications}, 77(12):15573--15601, 2018.

\bibitem{gong2021poseaug}
Kehong Gong, Jianfeng Zhang, and Jiashi Feng.
\newblock Poseaug: A differentiable pose augmentation framework for 3d human
  pose estimation.
\newblock In {\em {CVPR}}, 2021.

\bibitem{Hartley2003MVG}
Richard Hartley and Andrew Zisserman.
\newblock {\em Multiple View Geometry in Computer Vision}.
\newblock Cambridge University Press, New York, NY, USA, 2 edition, 2003.

\bibitem{he2016deep}
Kaiming He, Xiangyu Zhang, Shaoqing Ren, and Jian Sun.
\newblock Deep residual learning for image recognition.
\newblock In {\em {CVPR}}, 2016.

\bibitem{he2020epipolar}
Yihui He, Rui Yan, Katerina Fragkiadaki, and Shoou-I Yu.
\newblock Epipolar transformers.
\newblock In {\em {CVPR}}, 2020.

\bibitem{huang2020end}
Congzhentao Huang, Shuai Jiang, Yang Li, Ziyue Zhang, Jason Traish, Chen Deng,
  Sam Ferguson, and Richard~Yi Da~Xu.
\newblock End-to-end dynamic matching network for multi-view multi-person 3d
  pose estimation.
\newblock In {\em {ECCV}}, 2020.

\bibitem{ionescu2014human3}
Catalin Ionescu, Dragos Papava, Vlad Olaru, and Cristian Sminchisescu.
\newblock Human3. 6m: Large scale datasets and predictive methods for 3d human
  sensing in natural environments.
\newblock {\em {IEEE Trans. on Pattern Analysis and Machine Intelligence}},
  36(7):1325--1339, 2014.

\bibitem{iskakov2019learnable}
Karim Iskakov, Egor Burkov, Victor Lempitsky, and Yury Malkov.
\newblock Learnable triangulation of human pose.
\newblock In {\em {ICCV}}, 2019.

\bibitem{jiang2020coherent}
Wen Jiang, Nikos Kolotouros, Georgios Pavlakos, Xiaowei Zhou, and Kostas
  Daniilidis.
\newblock Coherent reconstruction of multiple humans from a single image.
\newblock In {\em {CVPR}}, 2020.

\bibitem{jiang2021transgan}
Yifan Jiang, Shiyu Chang, and Zhangyang Wang.
\newblock Transgan: Two transformers can make one strong gan.
\newblock {\em arXiv}, 2021.

\bibitem{joo2015panoptic}
Hanbyul Joo, Hao Liu, Lei Tan, Lin Gui, Bart Nabbe, Iain Matthews, Takeo
  Kanade, Shohei Nobuhara, and Yaser Sheikh.
\newblock Panoptic studio: A massively multiview system for social motion
  capture.
\newblock In {\em {ICCV}}, 2015.

\bibitem{joo2017panoptic}
Hanbyul Joo, Tomas Simon, Xulong Li, Hao Liu, Lei Tan, Lin Gui, Sean Banerjee,
  Timothy Godisart, Bart Nabbe, Iain Matthews, et~al.
\newblock Panoptic studio: A massively multiview system for social interaction
  capture.
\newblock {\em IEEE transactions on pattern analysis and machine intelligence},
  41(1):190--204, 2017.

\bibitem{kadkhodamohammadi2021generalizable}
Abdolrahim Kadkhodamohammadi and Nicolas Padoy.
\newblock A generalizable approach for multi-view 3d human pose regression.
\newblock {\em Machine Vision and Applications}, 32(1):1--14, 2021.

\bibitem{hmrKanazawa17}
Angjoo Kanazawa, Michael~J. Black, David~W. Jacobs, and Jitendra Malik.
\newblock End-to-end recovery of human shape and pose.
\newblock In {\em {CVPR}}, 2018.

\bibitem{kingma2014adam}
Diederik~P Kingma and Jimmy Ba.
\newblock Adam: A method for stochastic optimization.
\newblock In {\em {ICCV}}, 2015.

\bibitem{kreiss2019pifpaf}
Sven Kreiss, Lorenzo Bertoni, and Alexandre Alahi.
\newblock Pifpaf: Composite fields for human pose estimation.
\newblock In {\em {CVPR}}, 2019.

\bibitem{kuhn1955hungarian}
Harold~W Kuhn.
\newblock The hungarian method for the assignment problem.
\newblock {\em Naval research logistics quarterly}, 2(1-2):83--97, 1955.

\bibitem{lin2021multi}
Jiahao Lin and Gim~Hee Lee.
\newblock Multi-view multi-person 3d pose estimation with plane sweep stereo.
\newblock In {\em {CVPR}}, 2021.

\bibitem{lin2017focal}
Tsung-Yi Lin, Priya Goyal, Ross Girshick, Kaiming He, and Piotr Doll{\'a}r.
\newblock Focal loss for dense object detection.
\newblock In {\em {ICCV}}, 2017.

\bibitem{loper2015smpl}
Matthew Loper, Naureen Mahmood, Javier Romero, Gerard Pons-Moll, and Michael~J
  Black.
\newblock Smpl: A skinned multi-person linear model.
\newblock {\em ACM transactions on graphics (TOG)}, 34(6):1--16, 2015.

\bibitem{martinez2017simple}
Julieta Martinez, Rayat Hossain, Javier Romero, and James~J Little.
\newblock A simple yet effective baseline for 3d human pose estimation.
\newblock In {\em {ICCV}}, 2017.

\bibitem{mehta2017vnect}
Dushyant Mehta, Srinath Sridhar, Oleksandr Sotnychenko, Helge Rhodin, Mohammad
  Shafiei, Hans-Peter Seidel, Weipeng Xu, Dan Casas, and Christian Theobalt.
\newblock Vnect: Real-time 3d human pose estimation with a single rgb camera.
\newblock {\em {ACM Trans. on Graphics}}, 36(4):44, 2017.

\bibitem{nie2019spm}
Xuecheng Nie, Jianfeng Zhang, Shuicheng Yan, and Jiashi Feng.
\newblock Single-stage multi-person pose machines.
\newblock In {\em {ICCV}}, 2019.

\bibitem{papandreou2018personlab}
George Papandreou, Tyler Zhu, Liang-Chieh Chen, Spyros Gidaris, Jonathan
  Tompson, and Kevin Murphy.
\newblock Personlab: Person pose estimation and instance segmentation with a
  bottom-up, part-based, geometric embedding model.
\newblock In {\em {ECCV}}, 2018.

\bibitem{paszke2017automatic}
Adam Paszke, Sam Gross, Soumith Chintala, Gregory Chanan, Edward Yang, Zachary
  DeVito, Zeming Lin, Alban Desmaison, Luca Antiga, and Adam Lerer.
\newblock Automatic differentiation in pytorch.
\newblock In {\em NeurIPSw}, 2017.

\bibitem{pavlakos2017harvesting}
Georgios Pavlakos, Xiaowei Zhou, Konstantinos~G Derpanis, and Kostas
  Daniilidis.
\newblock Harvesting multiple views for marker-less 3d human pose annotations.
\newblock In {\em {CVPR}}, 2017.

\bibitem{popa2017deep}
Alin-Ionut Popa, Mihai Zanfir, and Cristian Sminchisescu.
\newblock Deep multitask architecture for integrated 2d and 3d human sensing.
\newblock In {\em {CVPR}}, 2017.

\bibitem{qiu2019cross}
Haibo Qiu, Chunyu Wang, Jingdong Wang, Naiyan Wang, and Wenjun Zeng.
\newblock Cross view fusion for 3d human pose estimation.
\newblock In {\em {ICCV}}, 2019.

\bibitem{remelli2020lightweight}
Edoardo Remelli, Shangchen Han, Sina Honari, Pascal Fua, and Robert Wang.
\newblock Lightweight multi-view 3d pose estimation through camera-disentangled
  representation.
\newblock In {\em {CVPR}}, 2020.

\bibitem{sun2018integral}
Xiao Sun, Bin Xiao, Fangyin Wei, Shuang Liang, and Yichen Wei.
\newblock Integral human pose regression.
\newblock In {\em {ECCV}}, 2018.

\bibitem{sutskever2014sequence}
Ilya Sutskever, Oriol Vinyals, and Quoc~V Le.
\newblock Sequence to sequence learning with neural networks.
\newblock {\em arXiv}, 2014.

\bibitem{Tu2020}
Hanyue Tu, Chunyu Wang, and Wenjun Zeng.
\newblock Voxelpose: Towards multi-camera 3d human pose estimation in wild
  environment.
\newblock In {\em {ECCV}}, 2020.

\bibitem{vaswani2017attention}
Ashish Vaswani, Noam Shazeer, Niki Parmar, Jakob Uszkoreit, Llion Jones,
  Aidan~N Gomez, Lukasz Kaiser, and Illia Polosukhin.
\newblock Attention is all you need.
\newblock {\em arXiv}, 2017.

\bibitem{wang2020end}
Yuqing Wang, Zhaoliang Xu, Xinlong Wang, Chunhua Shen, Baoshan Cheng, Hao Shen,
  and Huaxia Xia.
\newblock End-to-end video instance segmentation with transformers.
\newblock {\em arXiv}, 2020.

\bibitem{wu2019pay}
Felix Wu, Angela Fan, Alexei Baevski, Yann~N Dauphin, and Michael Auli.
\newblock Pay less attention with lightweight and dynamic convolutions.
\newblock {\em arXiv}, 2019.

\bibitem{wu2020lite}
Zhanghao Wu, Zhijian Liu, Ji~Lin, Yujun Lin, and Song Han.
\newblock Lite transformer with long-short range attention.
\newblock {\em arXiv}, 2020.

\bibitem{xiao2018simple}
Bin Xiao, Haiping Wu, and Yichen Wei.
\newblock Simple baselines for human pose estimation and tracking.
\newblock In {\em {ECCV}}, 2018.

\bibitem{zhang2020inference}
Jianfeng Zhang, Xuecheng Nie, and Jiashi Feng.
\newblock Inference stage optimization for cross-scenario 3d human pose
  estimation.
\newblock In {\em {NeurIPS}}, 2020.

\bibitem{zhang2021bmp}
Jianfeng Zhang, Dongdong Yu, Jun~Hao Liew, Xuecheng Nie, and Jiashi Feng.
\newblock Body meshes as points.
\newblock In {\em {CVPR}}, 2021.

\bibitem{zhao2020exploring}
Hengshuang Zhao, Jiaya Jia, and Vladlen Koltun.
\newblock Exploring self-attention for image recognition.
\newblock In {\em {CVPR}}, 2020.

\bibitem{zhou2017towards}
Xingyi Zhou, Qixing Huang, Xiao Sun, Xiangyang Xue, and Yichen Wei.
\newblock Towards 3d human pose estimation in the wild: a weakly-supervised
  approach.
\newblock In {\em {ICCV}}, 2017.

\bibitem{zhu2019deformable}
Xizhou Zhu, Han Hu, Stephen Lin, and Jifeng Dai.
\newblock Deformable convnets v2: More deformable, better results.
\newblock In {\em {CVPR}}, 2019.

\bibitem{zhu2020deformabledetr}
Xizhou Zhu, Weijie Su, Lewei Lu, Bin Li, Xiaogang Wang, and Jifeng Dai.
\newblock Deformable detr: Deformable transformers for end-to-end object
  detection.
\newblock In {\em {ICCV}}, 2020.

\end{thebibliography}
}

\clearpage
\section*{More Implementation Details}
We use PyTorch~\cite{paszke2017automatic} to implement the proposed \textbf{M}ulti-\textbf{v}iew \textbf{P}ose transformer (MvP) model. 
Our MvP model is trained on 8 Nvidia RTX 2080 Ti GPUs, with a batch size of 1 per GPU and a total batch size of 8. We use the Adam optimizer~\cite{kingma2014adam} with an initial learning rate of 1e-4 and decrease the learning rate by a factor of 0.1 at 20 epochs during training. 
The hyper-parameter $\lambda$ for balancing confidence score and pose regression losses is set to 2.5. 
We use the image feature representations (256-d) from the de-convolution layer of the 2D pose estimator PoseResNet~\cite{xiao2018simple} for multi-view inputs. Additionally, we provide the code of MvP, including the implementation of model architecture, training and inference, in the folder of ``./mvp'' for better understanding our method.

\section*{Architecture Details}

\begin{figure}[h!]
	\centering
	\includegraphics[width=\linewidth]{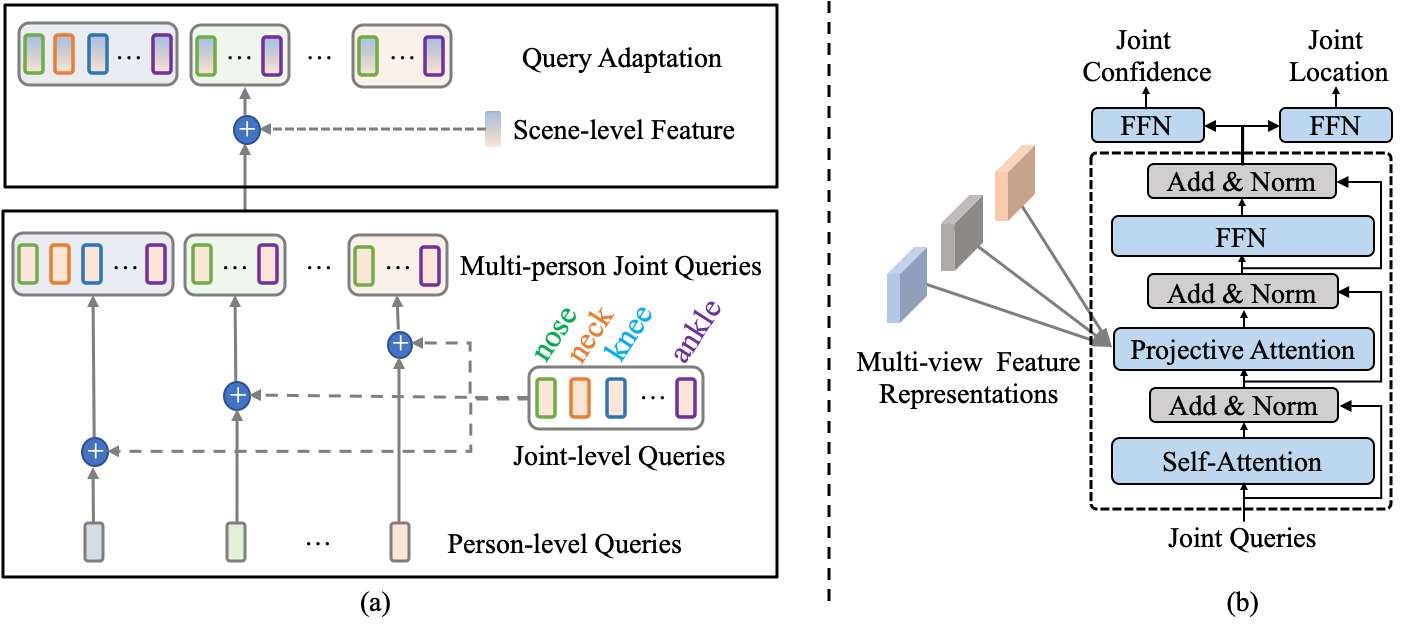}
	\caption{(a) Illustration of the proposed hierarchical query embedding and the input-dependent query adaptation schemes.  (b) Architecture of MvP's decoder layer. It consist of a self-attention, a projective attention and a feed-forward network (FFN) with residual connections. Add means addition and Norm means normalization. Best viewed in color.}
	\label{supp:fig1}
\end{figure}

\paragraph{Hierarchical Joint Query Embedding}
Fig.~\ref{supp:fig1} (a) illustrates our proposed 
hierarchical query embedding scheme. As shown in Eqn.~(1), 
each person-level query is added individually to the same set of    joint-level queries to obtain the per-person customized joint queries. 
This scheme shares the joint-level queries   across different persons and thus reduces the number of   parameters (the joint embeddings) to learn, 
and helps the model generalize better. The  generated per-person joint query embedding is further augmented by adding  the scene-level feature extracted from the input images.

\paragraph{Decoder Layer}
The decoder of MvP transformer consists of multiple decoder layers for regressing 3D joint locations progressively.
Fig.~\ref{supp:fig1} (b) demonstrates the detailed architecture of a decoder layer, which contains
a self-attention module to perform
pair-wise interaction between all the joints from multiple persons in the scene; a projective attention module
to selectively gather the complementary multi-view information; and a feed-forward network (FFN) to
predict the 3D joint locations and their confidence scores.

\section*{More Ablation Studies}

\paragraph{Replacing Camera Ray Directions with 2D Spatial Coordinates}
MvP encodes camera ray directions into the multi-view image feature representations via RayConv. We also compare with the simple positional embedding baseline that uses 2D coordinates as the positional information to embed, similar to the previous transformer-based models for vision tasks~\cite{carion2020end,dosovitskiy2020image}. 
Specifically, we replace the camera ray directions with 2D spatial coordinates of the input images in RayConv. Results are shown in Table~\ref{supp:exp:2dcoords}. We can observe using the 2D coordinates in RayConv results in much worse performance, \ie, 83.3 in AP$_{25}$ and 18.1 in MPJPE. This result demonstrates that using such view-agnostic 2D coordinates information cannot well encode multi-view geometrical information into the model; while using camera ray directions can effectively encode the positional information of each view in 3D space, thus leading to better performance.

\begin{table}[h]
    \centering
     \caption{Results of replacing camera ray directions with 2D coordinates in RayConv.}
	\begin{tabular}{cccc}
		\toprule
		Positional Input & AP$_{25}$ & AP$_{100}$ & MPJPE \\
		\midrule
		Camera Ray Directions  & 92.3 & 97.5 & 15.8 \\
		2D Spatial Coordinates  & 83.3 & 93.0 & 18.1 \\
		\bottomrule
	\end{tabular}
    \label{supp:exp:2dcoords}
\end{table}

\paragraph{Replacing Projective Attention with Dense Attention}
We further investigate the effectiveness of the proposed projective attention by  comparing it with  
the dense dot product attention, \ie, conducting attention densely over all spatial locations and camera views for multi-view information gathering.
Results are given in Table~\ref{supp:exp:vanilla_attn}. 
We observe MvP with the dense attention (MvP-Dense) delivers very poor performance (0.0 AP$_{25}$ and 114.5 MPJPE)
since it does not exploit any 3D geometries and thus is difficult to optimize.
Moreover, such dense dot product attention incurs significantly higher computation cost than the proposed projective attention\textemdash
MvP-Dense costs 31 G GPU memory, more than 5$\times$ larger than MvP with the projective attention, which only costs 6.1 G GPU memory.

\begin{table}[h]
    \centering
    \caption{
    Comparison between   the dense attention and the proposed projective attention. MvP-Dense means replacing the projective attention with the dense attention. We report GPU memory cost with a batch size of 1 during training.}
	\begin{tabular}{ccccc}
		\toprule
		Models & AP$_{25}$ & AP$_{100}$ & MPJPE & GPU Memory[G]\\
		\midrule
		MvP-Dense  & 0.0 & 16.1 & 114.5 & 31.0 \\
		MvP  & 92.3 & 97.5 & 15.8 & 6.1\\
		\bottomrule
	\end{tabular}
    \label{supp:exp:vanilla_attn}
\end{table}

\section*{More Results}

\paragraph{Quantitative Result}
We also evaluate our MvP model on the most widely used single-person dataset Human3.6M~\cite{ionescu2014human3} collected in an indoor environment. We follow the standard training and evaluation protocol~\cite{martinez2017simple,iskakov2019learnable,Tu2020} and use MPJPE as evaluation metric. Our MvP model achieves 18.6 MPJPE which is comparable to state-of-the-art approaches (18.6 \textit{v.s} 17.7 and 19.0)~\cite{iskakov2019learnable,Tu2020}.

\begin{figure}[t]
	\centering
	\includegraphics[width=0.95\linewidth]{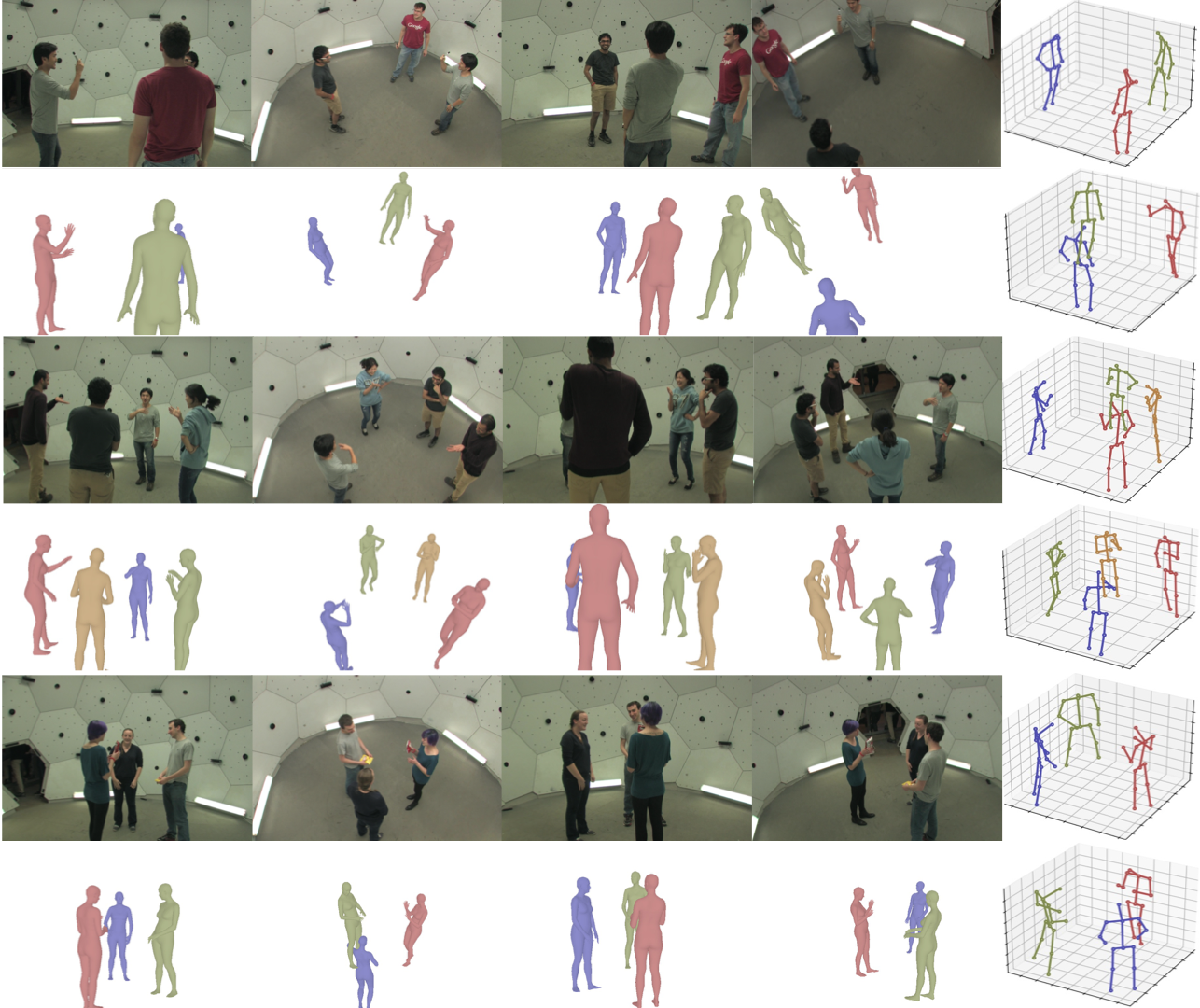}
	\caption{
    Example 3D pose estimations from Panoptic dataset. The left four columns show the multi-view inputs and the corresponding body mesh estimations from MvP. The rightmost column shows the estimated 3D poses from two different views.
    Best viewed in color. 
	}
	\label{supp:panoptic}
\end{figure}

\begin{figure}[h!]
	\centering
	\includegraphics[width=0.95\linewidth]{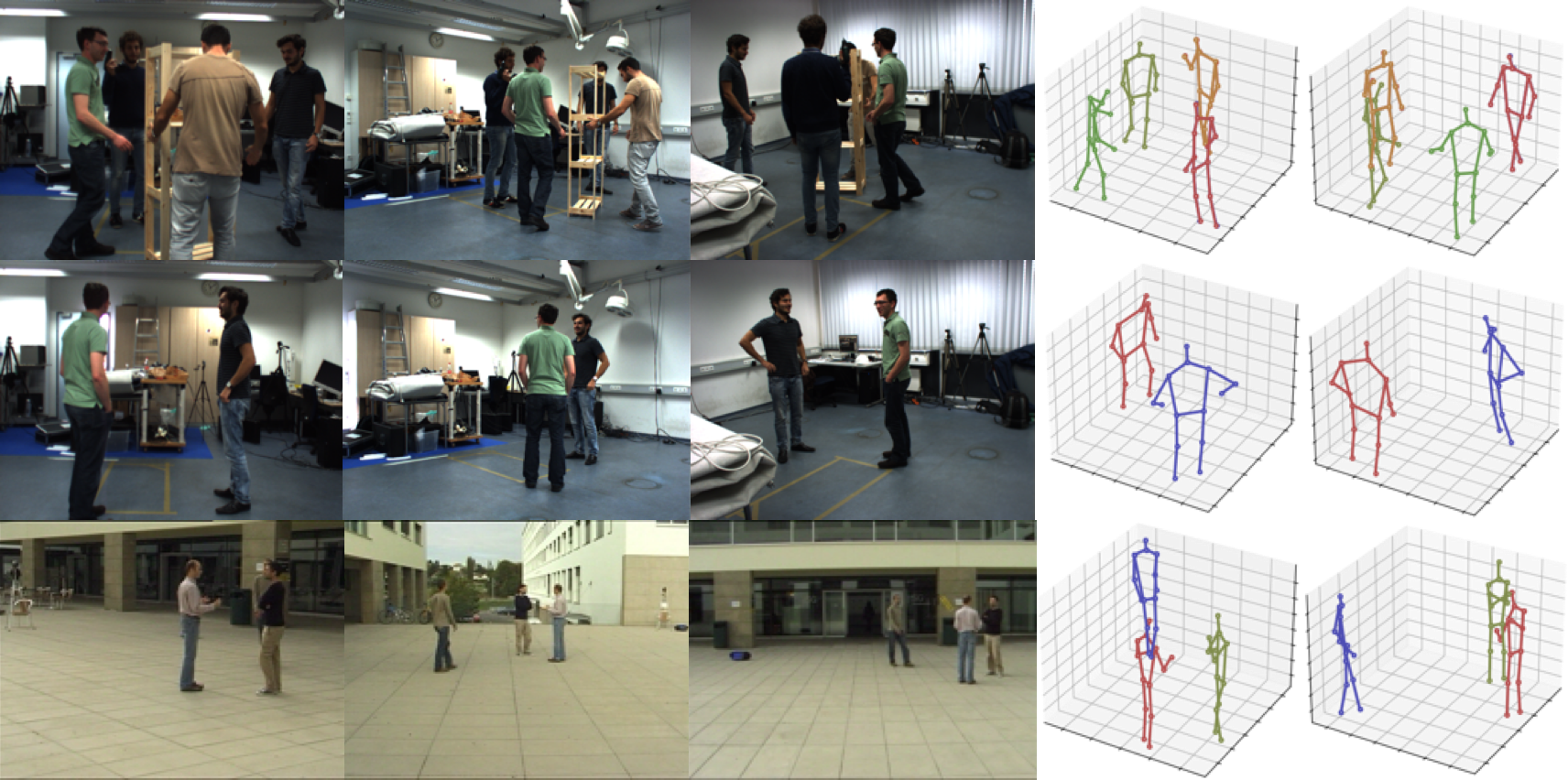}
	\caption{
    Example results of 3D pose estimation from MvP on the  Shelf (the 1st and 2nd examples) and Campus (the 3rd example) datasets. The left three columns show the multi-view inputs. The rightmost column shows the estimated 3D poses from two different views.
    Best viewed in color. 
	}
	\label{supp:shelf_and_campus}
\end{figure}

\paragraph{Qualitative Result}
Here we present more qualitative results of MvP on Panoptic~\cite{joo2017panoptic} (Fig.~\ref{supp:panoptic}), Shelf and Campus~\cite{belagiannis20143d} (Fig.~\ref{supp:shelf_and_campus}) datasets. From Fig~\ref{supp:panoptic} we can observe that MvP can produce satisfactory 3D pose and body mesh estimations even in case of strong pose deformations (the 1st example) and large occlusion (the 2nd and 3rd examples). Moreover, the performance of MvP is robust even in the challenging crowded scenario, as shown in the 1st example in Fig.~\ref{supp:shelf_and_campus}. 

\end{document}